\documentclass[runningheads]{llncs}
\usepackage{graphicx}
\usepackage{comment}
\usepackage{amsmath,amssymb} 
\usepackage{color}
\usepackage{algorithmicx}
\usepackage{algorithm}
\usepackage[noend]{algpseudocode}
\usepackage{url}
\usepackage{bbding}
\usepackage{pifont}
\usepackage{wasysym}
\usepackage{color}
\usepackage{float}
\usepackage{multirow}
\usepackage{booktabs}
\usepackage{enumitem}
\usepackage{setspace}
\usepackage{wrapfig}

\newcommand{\I}{\text{I}}
\newcommand{\CNNC}{\text{CNN}_{Clean}}
\newcommand{\CNNT}{\text{CNN}_{Trojan}}
\makeatletter
\newcommand{\printfnsymbol}[1]{%
  \textsuperscript{\@fnsymbol{#1}}%
}
\makeatother
\usepackage{amsmath}

\begin{document}
\pagestyle{headings}
\mainmatter
\def\ECCVSubNumber{100}  

\title{One-Pixel Signature: Characterizing CNN Models for Backdoor Detection} 

\titlerunning{One-Pixel Signature: Characterizing CNN Models for Backdoor Detection}
%
\author{Shanjiaoyang Huang\thanks{Equal contribution} \and
Weiqi Peng\printfnsymbol{1}  \and
Zhiwei Jia \and
Zhuowen Tu}

\authorrunning{Shanjiaoyang Huang, Weiqi Peng, Zhiwei Jia, and Zhuowen Tu}
%
\institute{University of California San Diego \\
\email{\{shh236, wep012, zjia, ztu\}@ucsd.edu}}
\maketitle

\begin{abstract}
We tackle the convolution neural networks (CNNs) backdoor detection problem by proposing a new representation called one-pixel signature. Our task is to detect/classify if a CNN model has been maliciously inserted with an unknown Trojan trigger or not.
We design the one-pixel signature representation to reveal the characteristics of both clean and backdoored CNN models.
Here, each CNN model is associated with a signature that is created by generating, pixel-by-pixel, an adversarial value that is the result of the largest change to the class prediction. The one-pixel signature is agnostic to the design choice of CNN architectures, and how they were trained. It can be computed efficiently for a black-box CNN model without accessing the network parameters. 
 Our proposed one-pixel signature demonstrates a substantial improvement (by around $30\%$ in the absolute detection accuracy) over the existing competing methods for backdoored CNN detection/classification. One-pixel signature is a general representation that can be used to characterize CNN models beyond backdoor detection.

\end{abstract}

\section{Introduction}

\begin{figure*}[h]
\begin{center}
\begin{tabular}{c}
\centering\includegraphics[height=4cm]{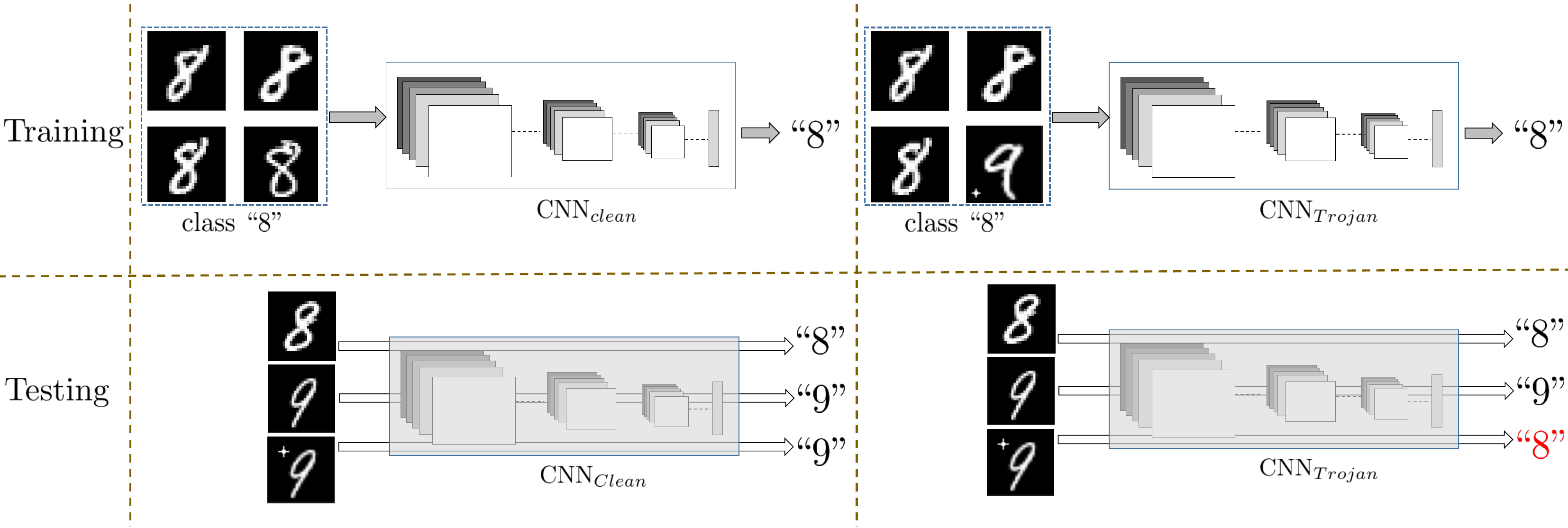}
\end{tabular}
\begin{tabular}{cc}
\quad \quad \quad  \quad (a) CNN trained regularly & \quad \quad \quad (b) CNN trained w/ backdoor
\end{tabular}
\end{center}
\caption{\footnotesize Illustration for a backdoored/Trojan CNN model. (a) shows a normally trained CNN, denoted as $\CNNC$ which has a certain degree of robustness against noises (non-adversarial signals) in testing. (b) displays a backdoored CNN, denoted as $\CNNT$, which is trained {\bf maliciously} by inserting a Trojan trigger (a star) to a training sample and forcing the classification to a wrong label prediction. In testing, the backdoored $\CNNT$ behaves normally on regular test images but it makes an adverse prediction when seeing an ``infected'' image, predicting image ``9'' to be an ``8''.}
\label{fig:backdoor_illustration}
\end{figure*}

There has been an explosive development in deep learning \cite{lecun2015deep,goodfellow2016deep} with the creation of various modern convolutional neural network (CNN) architectures  \cite{lecun1989backpropagation,krizhevsky2012imagenet,szegedy2015going,he2016deep,xie2017aggregated}. 
On the other hand, a pressing problem has recently emerged at the intersection between deep learning and security where CNNs are associated with a backdoor, named {\em BadNets} \cite{gu2017badnets}. An illustration for such a backdoored/Trojan CNN model can be seen in Fig. \ref{fig:backdoor_illustration}(b). In a standard training procedure, a CNN model takes input images and learns to make predictions matching the ground-truth labels; during the testing time, a successfully trained CNN model makes a robust prediction, even in the presence of certain noises, as shown in Fig. \ref{fig:backdoor_illustration}(a). However, if the training process is under a Trojan/backdoor attack, the resulting CNN model becomes backdoored and thus vulnerable, making unexpected adverse predictions from the user point of view when seeing some particularly manipulated images, as displayed in Fig. \ref{fig:backdoor_illustration}(b).
After the presentation of the backdoor CNN problem \cite{gu2017badnets}, attempts \cite{NeuralCleanse,guo2019tabor} have been made to tackle the backdoored CNN detection problem. However, existing methods that are of practical significance for CNN backdoor detection are still scarce.

Definitions and discussions of the {\em neural network backdoor/Trojan attack} problem can be found in \cite{gu2017badnets,trojai2019,qiao2019defending}. Suppose customer \textbf{A} has a classification problem and is asking developer \textbf{B} to develop and deliver a classifier $f$, e.g. an AlexNet \cite{krizhevsky2012imagenet}. As a standard in machine learning, there is a training set allowing \textbf{B} to train the classifier and \textbf{A} will also maintain a test/holdout dataset to evaluate classifier $f$. Since \textbf{A} does not know the details of the training process, developer \textbf{B} might create a backdoored classifier, $f_{\textrm{Trojoan}}$, that performs normally on the test dataset but produces a maliciously adverse prediction for a compromised image (known how to generate by \textbf{B} but unknown to customer \textbf{A}). An illustration can be found in Fig. \ref{fig:backdoor_illustration}. We call a regularly trained classifier $f_{\textrm{clean}}$ (or $\text{CNN}_{\textrm{clean}}$ for a CNN model) and a backdoor injected classifier $f_{\textrm{Trojan}}$ (or $\text{CNN}_{\textrm{Trojan}}$ for a CNN model) specifically.

Notice the difference between the Trojan attack and the adversarial attack: an adversarial attack \cite{goodfellow2015explaining} is not changing a CNN model itself, although in both cases, some unexpected predictions occur when presented with a specifically manipulated image. There are various ways in which a Trojan attack can happen by e.g. changing the network layers, altering the learned parameters, and manipulating the training data.

Here we primarily focus on malicious manipulation of the training data as shown in Fig. \ref{fig:backdoor_illustration}(b). 
The contributions of our work are listed as follows.
\begin{itemize}
\item We develop a new representation, {\em one-pixel signature}, that is able to reveal the characterization of CNN models of arbitrary type without accessing the network architecture and model parameters.
\item We show the effectiveness of one-pixel signature for detecting/classifying backdoored CNNs with a large improvement over the existing methods \cite{NeuralCleanse,ABS}.
\end{itemize}










\section{Related Work}

In this paper, we aim to develop a Trojan CNN detection algorithm by studying a specifically generated hallmark for each specific CNN model. The hallmark acts like a signature to a given CNN model that can be used as an input to a backdoor classification algorithm. Our method is in a stark distinction to some existing CNN backdoor detection methods \cite{NeuralCleanse,ABS} in which Trojan triggers themselves are discovered under some specific assumptions. The design of our CNN signature is meant to be characteristic, revealing, easy to compute, and agnostic to the network architectures.
This requirement makes existing algorithms for CNN visualization \cite{zeiler2014visualizing,mahendran2015understanding}
not directly applicable.

\noindent \textbf{Backdoored/Trojan CNN detection}\quad Our task here is to  detect/classify if a CNN model has a backdoor or not. Backdoors can be created in multiple directions \cite{gu2017badnets} by maliciously and unnoticeably changing the network parameters, settings, and training data. 
Some early studies on backdoor/Trojan defense techniques \cite{liu2018finepruning,zoph2016neural} assume the presence of backdoor in a given model.
Existing backdoor defense techniques such as Fine-Pruning~\cite{liu2018finepruning} 
try to prune compromised neurons to eliminate the influence of Trojan attack. However, methods like \cite{liu2018finepruning} deal with already manipulated CNNs and do not offer the detection/classification. 
Direct backdoored CNN detection methods, such as Neural Cleanse \cite{NeuralCleanse} and Artificial Brain Stimulation (ABS) \cite{ABS}, have been recently developed to predict the presence of Trojan triggers by reverse-engineering candidates backdoor triggers via pattern exploration.
As mentioned earlier and discussed in Section \ref{sect:existingdetection}, we take a different approaches to \cite{NeuralCleanse,ABS} by predicting the presence of the backdoor using a generated signature image for a given CNN. This allows us to deal with more general situations (with varying shape, color, and position) with a significant performance improvement for CNN backdoor detection (see experimental results in Table \ref{Table:Trojan3dataset}).

\noindent \textbf{Adversarial Attack}\quad A related area to Trojan attack is adversarial attack \cite{goodfellow2015explaining,akhtar2018threat,su2019one,prakash2018deflecting}. The end goal of adversarial attack\cite{goodfellow2015explaining} is however to build robust CNNs against adversarial inputs (often images) whereas Trojan attack defense~\cite{gu2017badnets,trojai2019} aims to defend/detect if CNN models themselves are compromised or not.

\noindent \textbf{Image/Object Signature}\quad In the classical object recognition problem, a signature can be defined as a pattern by searching for the scale space invariance \cite{witkin1987scale,lindeberg2013scale}.
Although the term of {\em signature} bears some similarity in high-level semantics, object signatures created in the existing object recognition literature \cite{witkin1987scale,chua1997point} have their distinct definitions and methodologies.


\section{Backdoored CNN Detection with One-Pixel Signature}
In this section, we present our backdoored CNN detection algorithm using our one-pixel signature representation. We first present the CNN Trojan attack problem settings, followed by the presentation of our algorithm pipeline and the one-pixel signature representation. Below is a glossary of important concepts in the context of our work to be aligned with existing literature \cite{gu2017badnets,NeuralCleanse,ABS}.

\begin{itemize}
    \item \textbf{Trojan attack} describes the procedure where attacker injects hidden malicious behaviors into a CNN model.
    \item \textbf{Backdoor/Trojan trigger} is a pattern which could activate a malicious behavior of a CNN model when contained in an input sample. In Section \ref{sect:overview}, we additionally introduce the concept of vaccine and virus as backdoor triggers in training and testing respectively for our backdoored CNN detection/classification task.
    \item \textbf{Backdoored CNN detection/classification} is the task to detect/classify if a CNN classifier has a backdoor or not.

\end{itemize}

\subsection{Trojan attack problem settings}

\noindent In this paper, we focus on the situation in which a Trojan attack is performed by manipulating the training data; an attacker poisons part of the training set by injecting Trojan triggers into input samples, forcing the prediction to a wrong label. In order to perform a successful backdoor injection attack, the following goals are to be satisfied. 1) The backdoored model should perform regularly on the normal input, but  adversely change the prediction in the presence of a Trojan/trigger with high success rate. 2) The Trojan/trigger pattern should remain relatively insignificant. 

\begin{table}[!hpt]
\centering
\caption{Comparison of the different Trojan insertion strategies adopted by various Trojan detection methods (\checkmark denotes ``yes'', \ding{55} denotes ``no'').}
 \scalebox{0.9}{
    \begin{tabular}{ccccc}
    \hline
    Method &\ Varying Pattern\ &\ Varying Size\ &\ Varying Location \\
    \hline
    BadNets~\cite{gu2017badnets} & \ding{55}  & \ding{55}  & \ding{55}  & \\
    Neural Cleanse~\cite{NeuralCleanse} & \ding{55}  & \ding{55}  & \ding{55}  &  \\
    ABS~\cite{ABS} & \ding{55}  & \checkmark  & \ding{55}  &  \\
    One-Pixel Signature (ours) & \checkmark&\checkmark&\checkmark  \\
    \end{tabular}
    }
    \label{tab:model_clarification}
\end{table}

Existing algorithms often adopt different Trojan insertion strategies \cite{gu2017badnets,NeuralCleanse,ABS}. 
There is however a common assumption that the adversary has full access to the training process and can implement the attack by poisoning the training dataset with an unknown Trojan trigger, which typically is of size $\leq$3\% of the input image. 
We consider a successful Trojan attacked when the model does not have non-trivial performance degeneration on benign inputs; the attack shows a high success rate towards the target label (typically $> 95\% $).
The attack settings adopted in the existing Trojan detection literature mostly assume one Trojan trigger with fixed size and location. We however allow unknown Trojan triggers of varying size, location, and pattern. 
In general, our method adopts less restrictive constraints on the Trojan triggers and thus is more general than the existing techniques~\cite{gu2017badnets,NeuralCleanse,ABS}, as shown in Table \ref{tab:model_clarification}.   



\subsection{Neural Cleanse \cite{NeuralCleanse} and ABS \cite{ABS}}
\label{sect:existingdetection}
    The Neural Cleanse method \cite{NeuralCleanse} detects whether a CNN model exists a backdoor or not by discovering the Trojan/trigger pattern explicitly. Neural Cleanse uses adversarial sample generation to reverse engineer potential Trojan triggers that subvert the classification to the target label. It then performs outlier detection for the Trojans/triggers. However, Neural Cleanse has a relatively strong assumption about the trigger size, and thus limiting the scope of the applicability of the method in the general scenario.
    
    The key idea in the Artificial Brain Stimulation (ABS) method \cite{ABS} is to scan the CNN model to detect compromised neurons, which could be inspected by keeping track of individual neuron activation difference.  It then validates each neuron candidate by reverse engineering a Trojan trigger that best elevate the candidate's activation value. Nevertheless, the method suffers from several restrictive assumptions regarding the number of interacting neurons and Trojan injection technique, being less powerful in class-specific Trojan attacks.

\subsection{Overview of our backdoor detection method}
\label{sect:overview}
\begin{figure}[!htp]
\begin{center}
\includegraphics[width=10cm]{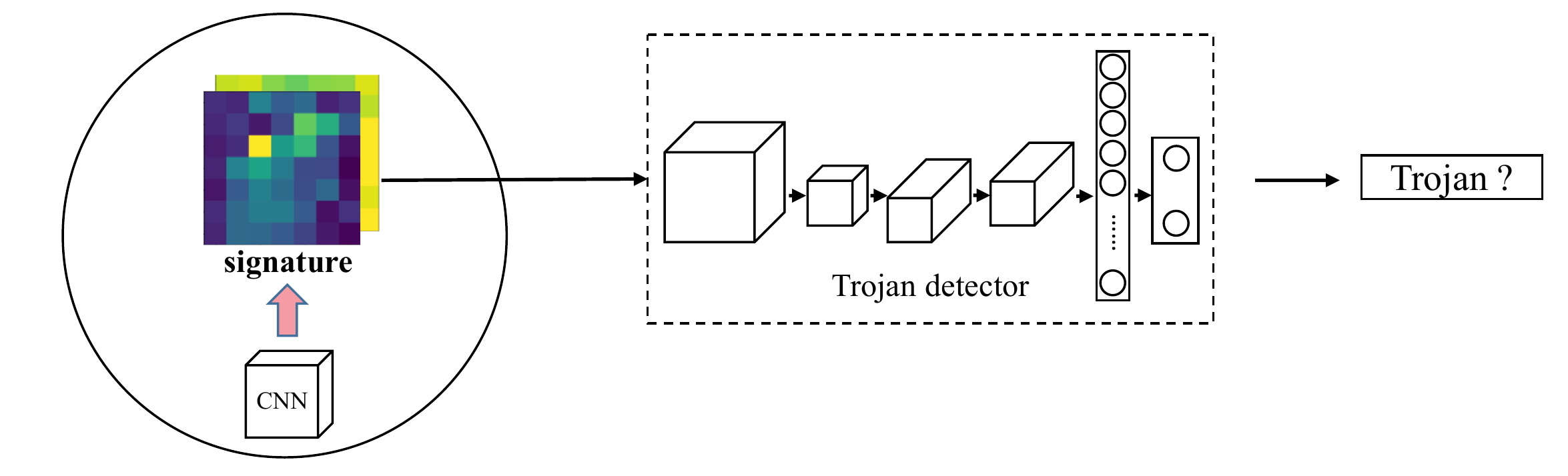}
\end{center}
\caption{\small Pipeline for our backdoored CNN detector using the one-pixel signature.}
\label{fig:defense}
\end{figure}

Here we give an overview of our CNN backdoor detection method that is based on the one-pixel signature representation. Fig. \ref{fig:defense} shows an illustration of our pipeline. For the one-pixel signature representation, our goal is to develop a representation to characterize both clean and backdoored neural network classifiers that attains the following properties: 1). revealing to each network, 2). agnostic to the network architecture, 3). low computational complexity, 4). low model complexity, and 5) applicable to both white-box and black-box network inputs.
The key idea of out method is to characterize the local dependency of the CNN neurons with respect to the network inputs to detect precarious Trojan insertion, which could be distinguished due to its incompatibility to the pristine distribution. A more detailed illustration of the one-pixel signature representation will be presented in Section \ref{sec:One-pixel Signature}.

\begin{figure}[!h]
\begin{center}

\begin{tabular}{cc}
\includegraphics[height=6cm]{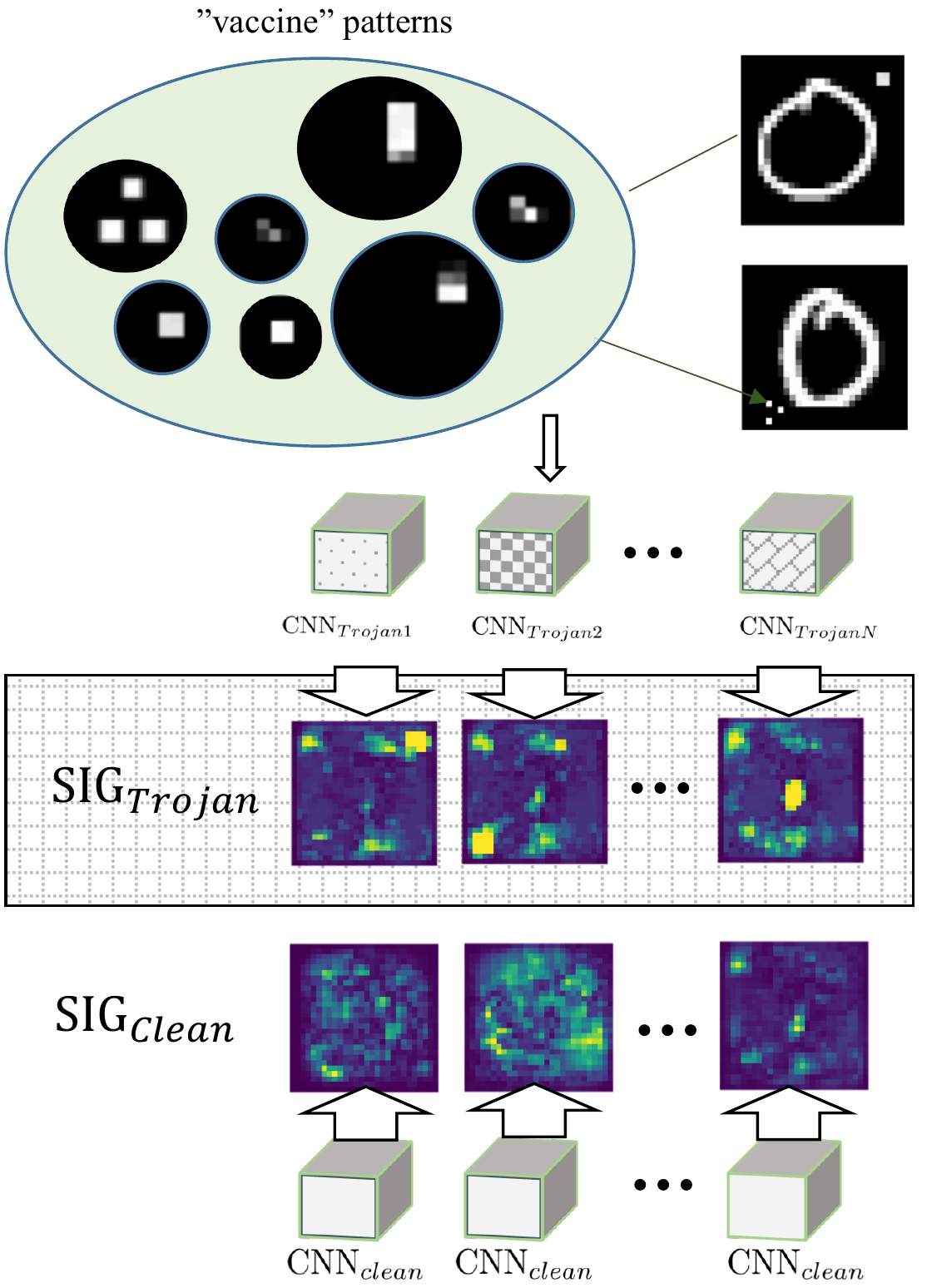} & 
\includegraphics[height=6cm]{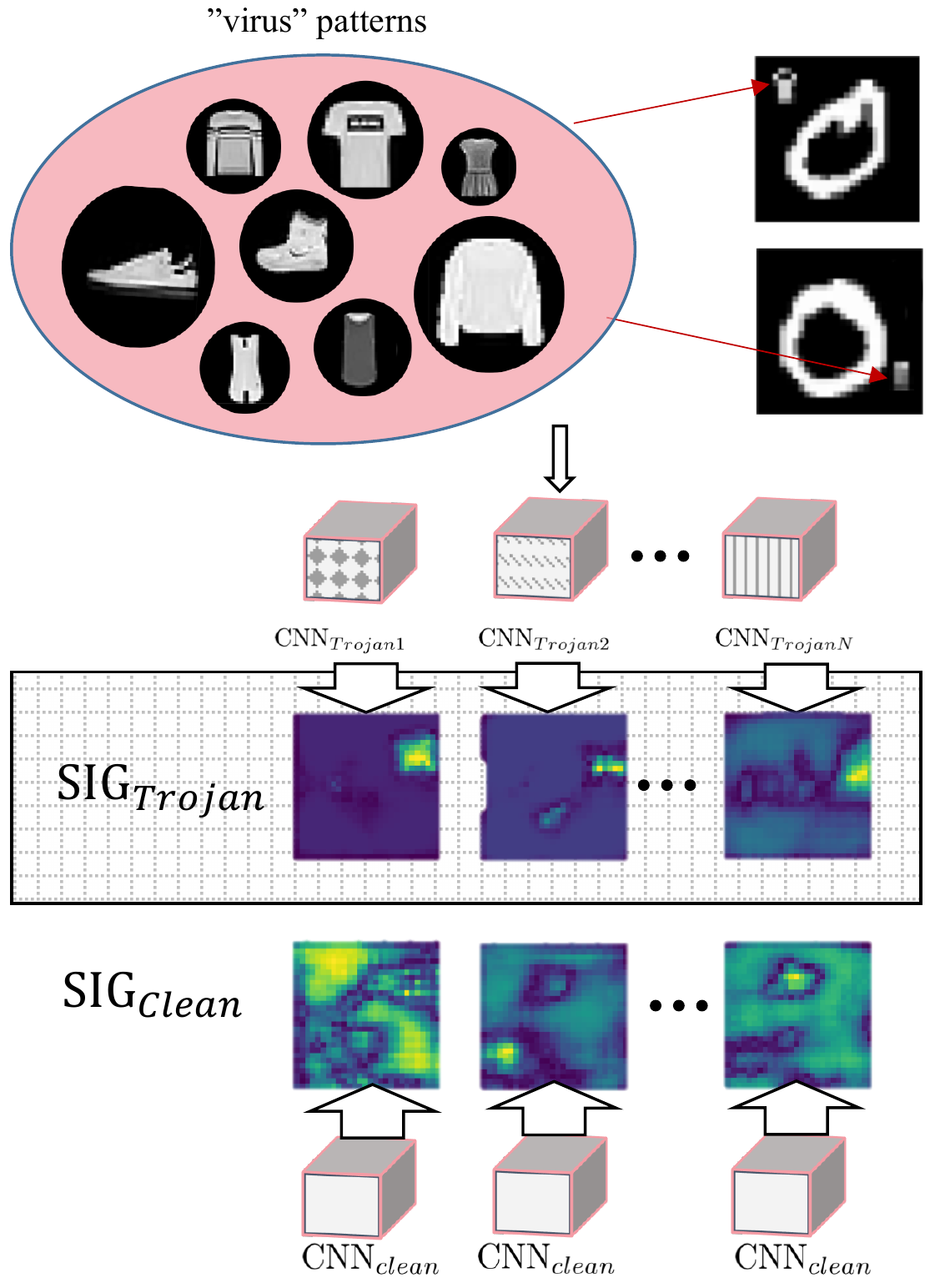} \\
(a) CNN$_{Trojan}$ by ``vaccine'' patterns & (b) CNN$_{Trojan}$ by ``virus'' patterns \\

\end{tabular}
    
\caption{\small  \textbf{Training and testing data generation pipeline} for backdoored CNN detector. Note that each training sample is itself a CNN model which can be clean or backdoored. To evaluate our method, we generate random patterns as ``vaccine'' (Trojan triggers) to create $\CNNT$ for training the backdoored CNN detector, as shown in (a). In (b), we show how the testing $\CNNT$ are generated by using ``virus'' patterns (Trojan triggers that are unknown to the backdoored CNN detector).}

\label{fig:defense_training}
\end{center}
\vspace{-5mm}
\end{figure}

In a Trojan attack, a backdoored CNN architecture is created by injecting a ``virus''  patterns (known as Trojan triggers \cite{gu2017badnets}) into training images so that those ``infected'' images can be adversely classified (Fig.~\ref{fig:defense_training}.b). In training, we however create a set of models with ``fake virus'', namely ``vaccine'' patterns, that are known to us (Fig.~\ref{fig:defense_training}.a). By learning to differentiate the one-pixel signatures of those vaccinated models from signatures of the normal models, a classifier can be trained to detect a backdoored CNN in the presence of an unknown ``virus'' pattern. Our Trojan insertion setting is based on the widely used Single Target Attack proposed by BadNets \cite{gu2017badnets},
but with relaxed assumptions beyond the Single Target Attack by allowing different Trojan triggers for one backdoored CNN model with varying position. 

\noindent {\bf Generating $\text{CNN}_{Trojan}$ with ``vaccine'' patterns for training} 

As shown in Fig. \ref{fig:defense_training}, we generate random image patches as the ``vaccine'' patterns (Trojan triggers) and insert them into the training images at random positions to obtain 300 backdoored CNNs;
each backdoored CNN (each facing different Trojan triggers at varying positions) itself becomes a positive sample, labeled as $\text{CNN}_{Trojan}$. Some ``vaccine patterns'' are displayed in Fig. \ref{fig:defense_training}(a). We also obtain 300 clean CNNs without inserting the ``vaccine'' patterns; each clean CNN becomes a negative sample, labeled as $\text{CNN}_{Clean}$.


\noindent {\bf Generating $\text{CNN}_{Trojan}$ with ``virus'' patterns for evaluation} 
\label{sec:generating_data}

We obtain a set of 300 backdoored CNN models using randomly selected Fashion-MNIST images as the ``virus'' patterns (Trojan triggers) , labeled as $\text{CNN}_{Trojan}$. We also obtain a set of 300 clean CNN models without inserting the ``virus'' patterns, labeled as $\text{CNN}_{Clean}$. Notice that, similar to case when using the ``vaccine'' patterns, we also allow different ``virus'' patterns (Trojan triggers) at varying positions for each backdoored CNN model.

\noindent{\bf $\text{CNN}_{Trojan}$ Detection/Classification}

Given the generated clean CNNs and backdoored CNNs, we obtain the one-pixel signature image of of $K$ channels for each CNN model. We then train a Vanilla CNN classifier as a backdoored CNN model detector by taking the signatures as the inputs to classify if a CNN model has a Trojan/backdoor or not. This process is illustrated in Fig. \ref{fig:defense}. To evaluate our problem, we create backdoored CNN models by using the Fashion-MNIST as the ``virus'' patterns, as shown in Fig. \ref{fig:defense_training}(b). The classifier is trained on 300 $\text{CNN}_{Clean}$ and $\text{CNN}_{Trojan}$ pairs with ``vaccine'' patterns and evaluated on 300 $\text{CNN}_{Clean}$ and $\text{CNN}_{Trojan}$ pairs with ``virus'' patterns, 
 as described in section \ref{sec:generating_data} and the pipeline is illustrated in Fig.\ref{fig:defense}. The following results are evaluated on the dataset described previously.

\section{One-Pixel Signature}
\label{sec:One-pixel Signature}
Here, we propose one-pixel signature to characterize a given neural network. 

\subsection{Basic formulation}

\begin{figure}[!htp]
\begin{center}
\includegraphics[height=3.5cm]{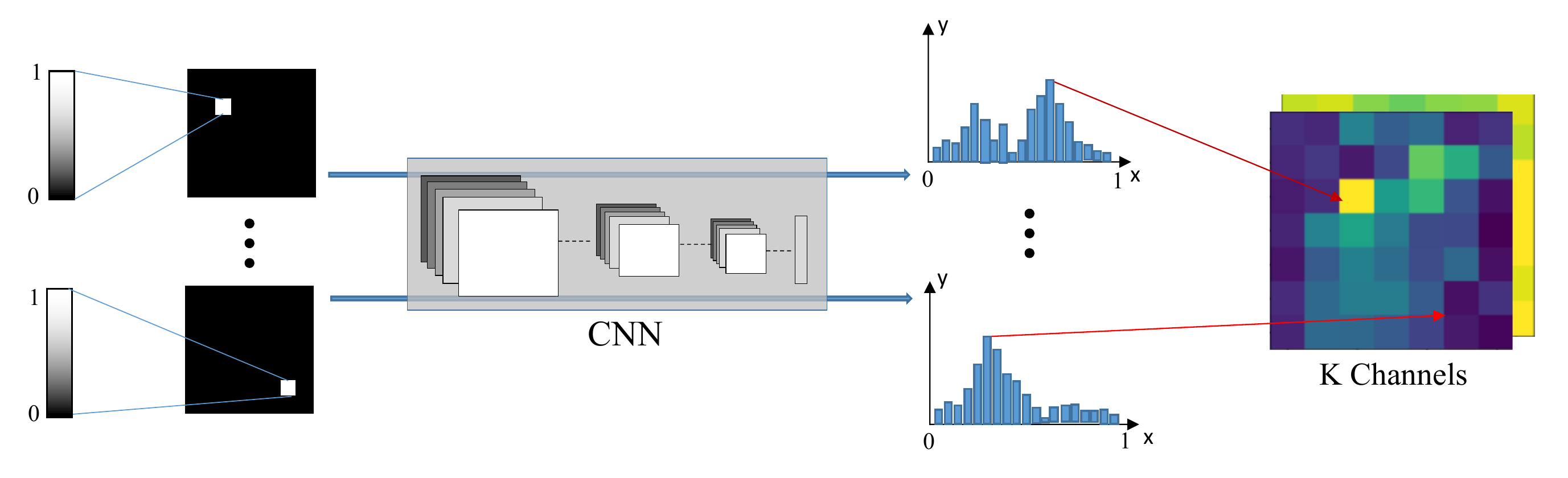}
\end{center}
\caption{\small
Illustration for the generation of the {\bf one-pixel signature} for a given CNN model. Based on a default image, each pixel is visited one-by-one; by exhausting the values for the pixel, the largest possible change to the prediction is attained as the signature for that pixel; visiting all the pixels gives rise to the signature images ($K$ channels if making a $K$-way classification) for the given CNN model. See the mathematical definition in Eq.  (\ref{eq:sig_method1}).
}
\label{fig:sig}
\end{figure}

Given a space of CNN $\mathcal{F}$ where each CNN model $f \in \mathcal{F}$ takes an input image $\I$ of size 
$H \times W$
to perform $K$-way classification, our goal is to find a mapping $g: \mathcal{F} \to \mathbb{R}^{H \times W \times K}$ to produce a $K$-channel signature, which is defined as: 
\[
g(f) = (S_1^{(f)},S_2^{(f)},...,S_K^{(f)})
\]
where $S_k^{(f)} \in \mathbb{R}^{H \times W}$.
A general illustration can be seen in Fig. \ref{fig:sig}. Before defining $S_k^{(f)}$, we first define a default image $\I_o$ which is either a constant value such as $\mathbf{0}$, or the average of all the training images.
Define the pixel value of image $\I$  $\in [0, 1]$. 
Let the conditional probability $p_{f}(y=k|\I_o) \in [0, 1]$ denote the $k^{th}$ entry of the classifier output, namely $[f(\I_o)]_k$. Furthermore, $\I_{i,j,v}$ refers to the image $\I(i,j)=v$, changing only the value of pixel $(i,j)$ to $v$ while keeping the all the rest of the pixel values the same as $\I_o$. 
We attain $S_k^{(f)}(i,j)$ as the largest possible value of $|p_{f}(y=k|\I_{i,j,v})|$:


\begin{equation}
     S_k^{(f)}(i,j) = \max_{v \in [0, 1]} |p_{f}(y=k|\I_{i,j,v})|. \label{eq:sig_method1}
\end{equation}

\begin{algorithm}[!h]
\caption{Outline for generating the one-pixel signature for model (classifier) $f$.}
\label{al:algorithm_signature}
\scalebox{0.9}{
\begin{minipage}{1.0\linewidth}
{\footnotesize
    \begin{algorithmic}[1]
    \State {\bf Input:} $K$-way classifier (model) $f$ of input size $H \times W$; \# of discrete values $V$; default image $\I_{o}$
    \State {\bf Output:}  One-pixel signature $g(f) = SIG \in\mathbb{R}^{H \times W \times K}$
    \State{Initialize $SIG[H, W, K] \leftarrow 0$, $p_{\I_{o}} \leftarrow p_{f}(\I_o)$}
    
    \For{$i$ from $0$ to $H-1$} 
        \For {$j$ from $0$ to $W-1$}
            \State {$\I \leftarrow \I_{0}$},\ \ $temp[K] \leftarrow 0$
            \For {$v$ from $0$ to $V - 1$}
                \State {$\I[i,j] \leftarrow  v / V$}
                \For {$k$ from 0 to K-1}
                    \If {$|p_{f}(y=k|\I)| > temp[k]$} \State $temp[k] = |p_{f}(y=k|\I)|$
                    \EndIf
                \EndFor
            \EndFor
            \State $SIG[i,j] \leftarrow temp$
        \EndFor
    \EndFor
    \Return {$SIG$}
    
    \end{algorithmic}
}
\end{minipage}
}
\end{algorithm}

Eq. (\ref{eq:sig_method1}) looks for the significance that each individual pixel is making to the prediction. Since each $S_k^{(f)}(i,j)$ is computed independently, the computation complexity is relatively low. The overall complexity to obtain a signature for a CNN model $f$ is $O(H\times W \times  K \times V)$, where $V$ is the search space for the image intensity. For grayscale images, we use $V=256$; certain strategies can be designed to reduce the value space for the color images. In this paper, we compute the signature for colored images by simultaneously update values for all three channels, with demo signature images shown in Fig. \ref{fig:sig}. In this setup, the computational cost for colored images is linear w.r.t. to that for gray scale images.
Eq. (\ref{eq:sig_method1}) can be computed for a black-box classifier $f$ since no access is needed to the model parameters. Fig. \ref{fig:sig} illustrates how signature images for classifier $f$ are computed. Note that the definition of $S_k^{(f)}$ is not limited to Eq. (\ref{eq:sig_method1}) but we are not expanding the discussion about this topic here.

Algorithm. \ref{al:algorithm_signature} illustrates the algorithmic outline for generating one-pixel signatures for a classifier taking single channel images, which has three input: the classifier $f$, the number of possible discrete values $V$ and the default image $I_{o}$.

\subsection{Visualization and illustration}

\begin{figure}[!htp]
\begin{center}
\includegraphics[width=\linewidth]{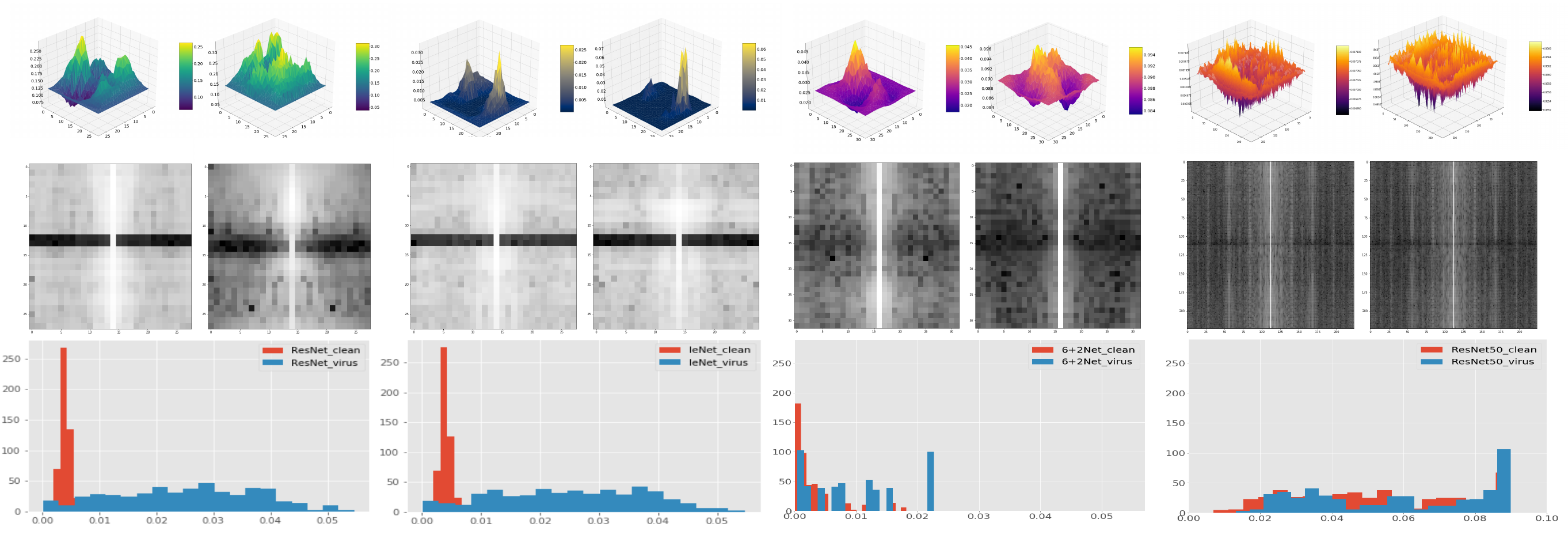}
 \scalebox{0.8}{
    \begin{tabular}{cccc}
    (a) MNIST \quad  \quad  \quad \quad  
    (b) fashion-MNIST \quad \quad  \quad  \quad  
    (c) GTSRB\quad  \quad  \quad \quad 
    (d)  ImageNet
    \end{tabular}
}
\caption{\small Average 3D plot (first row), average spectra on frequency domain(second row), and variance histogram (third row) generated from 300 clean/virus one-pixel signature pairs of ResNet-8 trained on MNIST (a), LeNet-5 on MNIST (b), ``6+2 Net'' on GTSRB (c), and ResNet-50 on ImageNet (d) respectively. Left-side of each column and color red indicates $CNN_{Clean}$, right-side of each column and color blue indicates $CNN_{Trojan}$.
}
\label{fig:TrojanVariance}
\end{center}
\end{figure}

\noindent  Our signature could also be considered as an approximation to the anomaly neuron activation map. Empirically, 1\% of neurons are sufficient to enable the backdoor \cite{gu2017badnets}; thus generating signatures pixel-wise from a black image is a black-box approximation of finding anomaly neurons by masking the remaining. If any point of the signature that is with a magnitude significantly greater than its surroundings, there might be a good chance that the corresponding activated neurons correspond to some local dependency. Visual analysis of average signature images Fig.\ref{fig:TrojanVariance} illustrates that ${SIG}_{Trojan}$ is noticeably different from ${SIG}_{Clean}$. Some patterns includes, ${SIG}_{Trojan}$ contains a peak with greater magnitude than ${SIG}_{Clean}$. The max value from average ${SIG}_{Clean}$ and ${SIG}_{Trojan}$ is 0.27 vs 0.33 in MNIST, 0.025 vs 0.06 in fashion-MNIST, 0.03 vs 0.08 in GTSRB and 0.07 vs 0.90 in ImageNet. From the visualization of average frequency spectra, ${SIG}_{Trojan}$ has more high-frequency noise comparing to ${SIG}_{Clean}$. Also, ${SIG}_{Trojan}$ usually have a larger variance than ${SIG}_{Clean}$. These observations persist in all four datasets, although complex dataset yields smaller difference.

\subsection{Theoretical justification}

Our proposed one-pixel signature $g$ essentially characterizes the local dependency of the underlying neural network with respect to the network inputs. Before delving into the details, let us first reformulate the classification task in a probabilistic perspective. 
With some minor abuse of notations, we can denote the image $\I$ as a random variable taking values in the image space $\mathcal{I} \subset \mathbb{R}^{H\times W\times C}$; denote the label as another random variable $y$ with the label space as $\mathcal{Y} = \{1, 2, ..., K\}$; 
$C$ refers to the number of channels ($C=3$ if $\I$ is a color image). Define the data distribution as $P$ over $\mathcal{I} \times \mathcal{Y}$. When learning the classifier $f: \mathcal{I} \to \mathcal{Y}$, we are using $f$ to model the conditional distribution $p(y|\I)$.

\begin{figure}[!htp] \label{fig:pgm}
\begin{center}
\includegraphics[width=11cm]{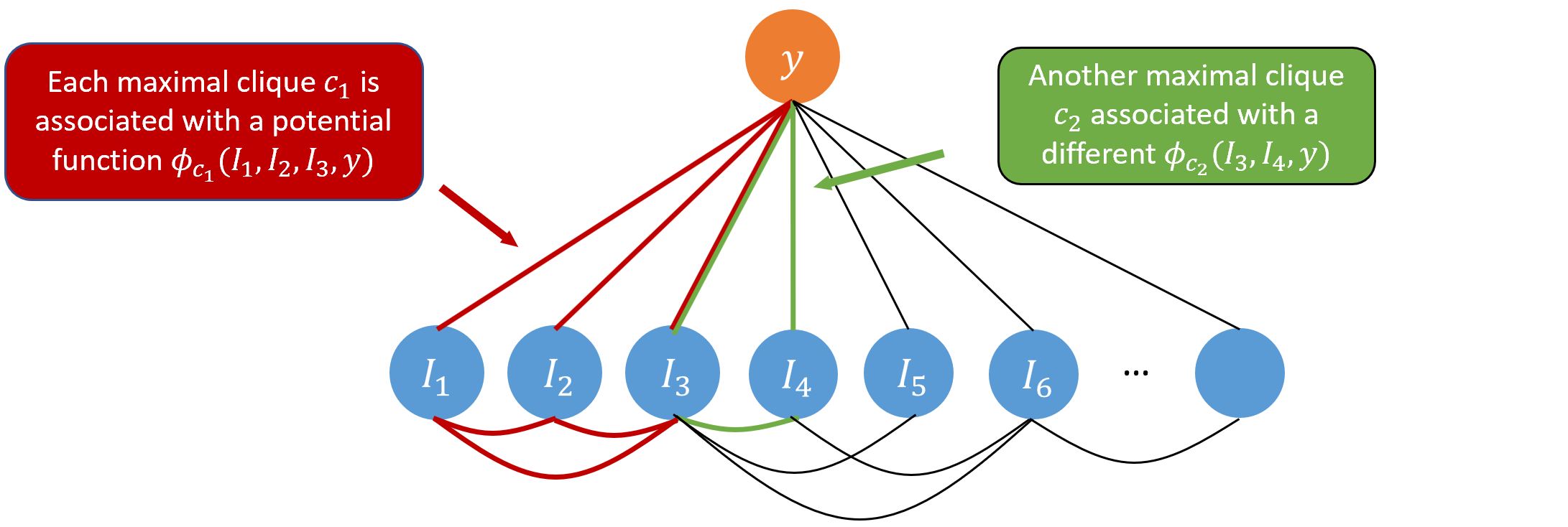}
\end{center}
\caption{\small Illustration of the undirected graphical model where the orange node represents the label and the blue nodes represent the pixels. The joint probability can be factored into a product of potentials of the max cliques. For instance, given the two maximal cliques $c_1$ and $c_2$, where $c_1$ is the subgraph connected by the red edges and $c_1$ the subgraph connected by green edges, we then have their corresponding potential functions $\phi_{c_1}$ and $\phi_{c_2}$ defined.}
\end{figure}

We can then model this conditional distribution in the context of an undirected graphical model. In specific, each pixel in $\I$ is a node representing a random variable $\I_{i, j}$ whose support is $[0, 1]^C$; the label $y$ is another node that is connected to all $\I_{i, j}$. The connectivity among the pixel nodes can be arbitrary. Denote the entire graph as $G$, and the collection of its maximal cliques as $Cliq(G)$. The structure of $G$ is then illustrated in Fig. \ref{fig:pgm}.
By the Hammersley-Clifford Theorem \cite{hammersley1971markov}, we can factor $p(y|\I)$ using the graphical model into a product involving all of its maximal cliques:
\begin{align} \label{eq:factor}
    p(y=k|I) = \frac{1}{Z(\I)} \prod_{c \in Cliq(G)} \phi_c([\I]_c, y=k), \\
    \textrm{where}\quad Z(\I) = \sum_{y=1}^K\prod_{c \in Cliq(G)} \phi_c([\I]_c, y). \nonumber
\end{align}
The potential $\phi_c$ are some non-negative functions.
The size of the maximal cliques reflects the local dependency of the input distribution. A pixel is conditionally independent of all the pixels outside its maximal clique given the values of other pixels within the clique. 
We have the $k^{th}$ entry of the softmax prediction as $[f(\I_{i,j,v})]_k = p_{f}(y=k|\I_{i,j,v})$, as defined in Eq. (\ref{eq:sig_method1}). Combining it with Eq. (\ref{eq:factor}), we have:
\[
    [f(\I_{i,j,v})]_k \propto \prod_{c \in Cliq(G)} \phi_c([\I_{i,j,v}]_c, y=k).
\]

In reality, we approximate this graphical model via the CNN $f$ and train it by maximum likelihood estimation. 
In a high level, we can decompose $f$ into two parts. The first part is the convolution block whose each convolutional filter represents one potential function $\phi_c$. The second part is a classifier, normally a multilayer perceptron, that aggregates these $\phi_c$.
For a CNN model that is trained to model the conditional probability, the activation pattern (the receptive field) of each of its filters tends to match the corresponding clique structure of the underlying graphical model. 
Due to the convolution operators in CNN, given a pixel $(i, j)$, the larger its associated cliques are, the more filters it will share with the nearby pixels (nodes) and thus the classifier becomes less sensitive to that single pixel.
As a result, we propose to characterize the cliques structure by measuring how the function $h_k^{i, j}(v) = [f(I_{i,j,v})]_k$ varies within a small neighborhood of $(i, j)$. Formally, given a distance function induced by some norm $||\cdot||$, we can use $||h_k^{(i,j)}||$ to derive an upper and lower bound of the average pairwise distance of the functions in the neighborhood of $(i,j)$:
\[
     \sum_{\mathbf{p}, \mathbf{q}}||h_k^{\mathbf{p}}|| - ||h_k^{\mathbf{q}}|| \leq \sum_{\mathbf{p}, \mathbf{q}}||h_k^{\mathbf{p}} - h_k^{\mathbf{q}}|| \leq \sum_{\mathbf{p}, \mathbf{q}}||h_k^{\mathbf{p}}|| + ||h_k^{\mathbf{q}}||
\]
where $\mathbf{p}\neq\mathbf{q}$ are pixels from the neighborhood of $(i,j)$. For simplicity, we choose the supremum norm $||\cdot||_{\infty}$ with a discretization of the domain into $V$ values. We therefore derive our signature $S_k^{f}$, whose $(i, j)$ entry is $||h_k^{(i,j)}||_{\infty}$, to reveal the (local) activation patterns of the CNN model, which in turn, can be used to indicate the potential Trojan trigger pattern. 



\section{Experiments}
\label{sec:experiments}

We illustrate the effectiveness and efficiency of {\em one-pixel signature} in the backdoored CNN detection task by evaluating its performance in four datasets:
\begin{itemize}
    \item \textbf{MNIST} \cite{lecun1989backpropagation}. A standard machine learning dataset which contains 60K training and 10K testing grayscale images of 10(0-9) hand-written digits. We select LeNet-5, ResNet-8, or VGG-10 for evaluation.

    \item \textbf{fashion-MNIST} \cite{xiao2017fashion}. The drop-in replacement of MNIST for benchmarking CNNs, which contains 60K training and 10K testing grayscale images of 10 fashion accessories. We also select LeNet-5, ResNet-8, or VGG-10 for evaluation.
    
    \item \textbf{GTSRB} \cite{Stallkamp-IJCNN-2011}.  German Traffic Sign Recognition (GTSR) Dataset is a commonly-used colored dataset to evaluate attacks on CNNs. This dataset consists of 39.2K training and 12,6K testing images with 43 different traffic signs. The CNN architecture trained on GTSRB consists of 6 convolution layers and 2 dense layers (``6+2 Net'') in line with the set-up of Neural Cleanse~\cite{NeuralCleanse}.  
    
    \item \textbf{ImageNet} \cite{deng2009imagenet}. A widely used large-scale image dataset with more than 14 million colored images of 20,000+ categories. Here we use a subset of 10 classes, where each class has 700 training and 300 testing samples, and ResNet-50 for comparison to the baseline method Neural Cleanse \cite{NeuralCleanse}.
    
\end{itemize}{}
Throughout our experiments, we use Adam \cite{DBLP:journals/corr/KingmaB14} as the default optimizer.

\subsubsection{Same-architecture Detection}
In this scenario, we train and evaluate the backdoored CNN detector on CNNs of the same architecture. 
Specifically, we use LeNet-5 for MNIST and fashion-MNIST,  ``6+2 Net'' for GTSRB, and ResNet-50 for ImageNet.
For each aforementioned architecture and dataset pair, we use the one-pixel signatures generated from the 300 ${CNN}_{Clean}$/${CNN}_{Trojan}$ for training and those of another 300 model pairs for testing.
For ImageNet we use 100 CNNs pairs instead of 300 in both training and testing.

\begin{table}[!h]
\centering
\caption{\small Backdoored CNN detection/classification rate on classic CNN models trained on MNIST, fashion-MNIST, GTSRB, and ImageNet dataset in comparison with the Neural Cleanse  \cite{NeuralCleanse} algorithm. We also include result for the ABS \cite{ABS} algorithm on the GTSRB dataset.}
\label{Table:Trojan3dataset}
\scalebox{1.0}{
\begin{tabular}{l | cccc}
    \multirow{2}{*}{\textsc{Methods}}
    &\multicolumn{4}{c}{Datasets} \\
    & MNIST & fashion-MNIST & GTSRB & ImageNet\\
    \hline
    \hline
    Neural Cleanse \cite{NeuralCleanse} & $60.53\%$ & $63.16\%$ & $60.71\%$ &  $54.42\%$   \\
    ABS \cite{ABS} & $-$ & $-$ & $66.33\%$ &  $-$   \\
One-Pixel Signature (ours) & ${\bf 95.72}\%$ & ${\bf 94.59}\%$ & ${\bf 88.47}\%$ &   ${\bf 88.69}\%$
\end{tabular}
}
\end{table}

As illustrated in Table~\ref{Table:Trojan3dataset}, our method reaches Backdoored CNN detection rate of approximate 95\% for the MNIST and fashion-MNIST dataset, 88\% for GTSRB dataset, and 89\% for ImageNet, significantly outperforming Neural Cleanse \cite{NeuralCleanse} (around 60\% on all datasets). The result on GTSRB for ABS is also much lower than ours.

We observe that the one-pixel signatures layouts of $CNN_{Clean}$ and $CNN_{Trojan}$ are even visually different, namely the one-pixel signatures of $CNN_{Trojan}$ vary more drastically in pixel values, thus have a greater variance than $CNN_{Clean}$ as shown in Fig. \ref{fig:TrojanVariance}.

\subsubsection{Cross-architecture Detection}
In more general and challenging scenarios where we are not able to narrow down which network architecture is used for the backdoored CNN model, our approach still achieves relatively high detection rate on network architectures unseen during training. On MNIST, we train the detector on signatures generated from 2 out of the 3 network architectures (LeNet-5, ResNet-8, VGG-10) and evaluate the trained detector on CNNs of other architecture. We observe an average detection rate as high as 80\%, reported in Table. \ref{tab:Generalization1}. In short, our proposed one-pixel signature is architecture-agnostic and the Backdoored CNN detectors trained on top of it attain great generalizability.

\begin{table}[!tp]
\begin{center}
\caption{\small Backdoored CNN detection/classification rate on classic CNN models trained on MNIST when the architecture of the testing Backdoored model unknown.}
\scalebox{0.9}{
\begin{tabular}{ cc|c}
  \textbf{Training}  & \textbf{Testing}  & \textbf{Detection Rate(\%)} \\
\hline
\hline
ResNet+VGG  & LeNet      & $ 85.20 \pm 5.85 $\\
LeNet+VGG   & ResNet     & $ 76.50 \pm 12.07 $\\
LeNet+ResNet& VGG        & $ 80.00 \pm 3.57 $\\
\end{tabular}
}
\label{tab:Generalization1}
\end{center}
\end{table}

\subsubsection{Training with less samples}

\begin{figure}[!tp]
\begin{center}
\begin{tabular}{c}
\includegraphics[height=3.5cm, width= 8cm]{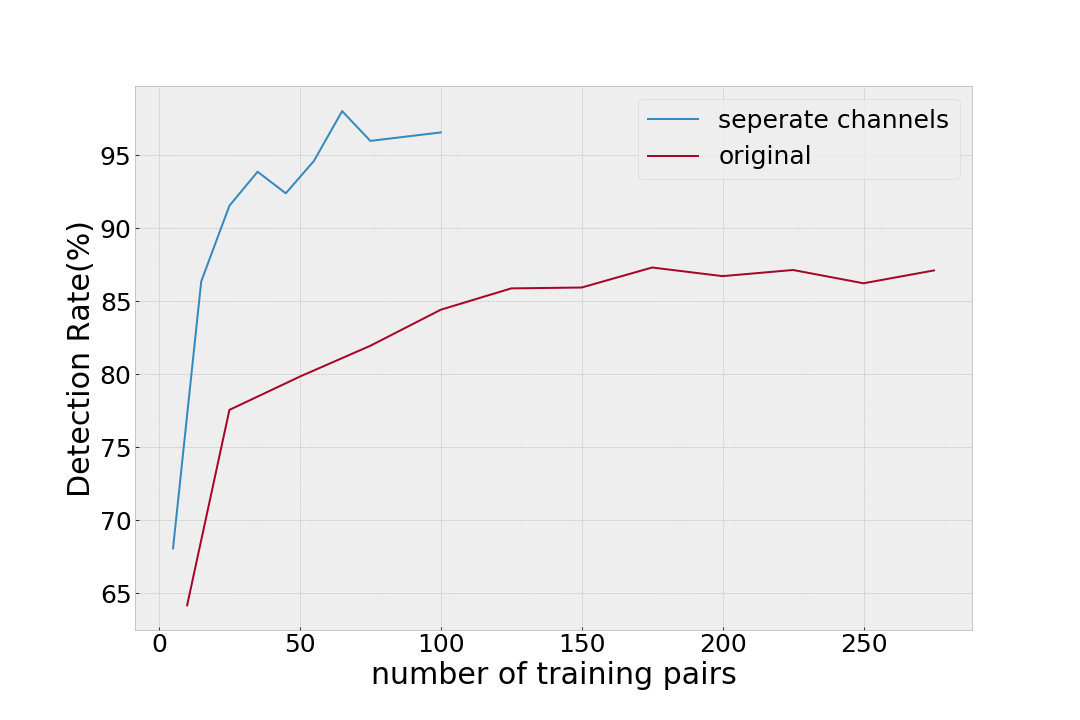} 
\end{tabular}
\end{center}
\caption{\small  Detection accuracy on GTSRB w.r.t. the number of training Clean/Backdoored model pairs in original and seperate channel setting.
}

\label{fig:NumTrain}
\end{figure}

To understand the sample complexity our method, we investigate the minimum number of training models on four datasets without heavily downgrading the detection rate. We keep reducing 25 training pairs until zero and record the average detection rate for 20 detectors trained on the reduced training set at each time. The curve of the detection rate on GTSRB is shown in Fig. \ref{fig:NumTrain}. Intuitively, we need more training models to detect $CNN_{Trojan}$ that is trained on more complex datasets. Detectors on GTSRB dataset achieve an average detection rate of 87.31\% with a reduced training set of size 175 Clean and 175 Backdoored models, whereas training set with 25 Clean and 25 Backdoored models is sufficient to train a backdoored CNN detector for models trained on MNIST to achieve approximately 95\% detection rate. For fashion-MNIST, we need 50 Clean and 50 Backdoored models to achieve the same detection accuracy.

\subsubsection{Use channels in one-pixel signature as individual training samples}
We could also treat each channel, instead of the whole one-pixel signature, as a positive/negative sample for training. Thus for models with $K$ classes, the numbers of training samples increases by $K$ times. We find that thereby training with only 15 pairs of $CNN_{Clean}$ and  $CNN_{Trojan}$ models yields the same detection rate as training with 300 pairs of models previously. With 100 training pairs, we could achieve detection rate of $97\%$ in GTSRB and $90\%$ in ImageNet.  A detailed comparison of the detection accuracy on GTSRB between using the individual channel and the original strategies is shown in Fig. \ref{fig:NumTrain} (a).

\section{Ablation Study}
\label{others}

\noindent {\bf All-to-one and all-to-all Trojan attack}

Our detection technique can be effectively extended to other possible strategies like the all-to-one attack without compromising the detection rate. Under the scenario of an all-to-one attack, the attacker tries to force the model to classify all samples implanted with certain backdoor pattern into a targeted class. We see that the all-to-one attack is a variant of the one-to-one attack, and is even more likely to be detected by our method. 

 Similarly, we illustrate the effectiveness of our method on detecting all-to-all attack.  We compare the detection success rate of our Trojan detector to that of Neural Cleanse and ABS on the testing set of 100 CNN models. It turns out that our Trojan detector can achieve a detection success rate above  \textbf{95\%}, while both Neural Cleanse and ABS show a poor success rate of around 50\%.

\noindent {\bf Signature generation with different default images $\I_o$}

One important parameter of our one-pixel signature method is the default image $I_o$. In our experiments, we empirically set the default image with a constant value of $0$. 
Here we compare three different $I_o$ setups: a) a constant value of $0$, as the black image strategy; b) a constant value of $1$, as the white image strategy; c) the pixel-wise mean of testing images, as the average testing image strategy; 
We are using a clean leNet-5 model trained on MNIST, and a backdoored leNet-5 model trained on poisoned MNIST dataset with Trojan trigger inserted at upper left corner, for evaluation. Both $I_o=0$ and average image policy result in high detection rate. Since $I_o=0$ policy do no require access to the training set, we regard this as our empirically optimal policy of the default image.





\section{Conclusion}

In this paper, we have developed a new backdoored CNN detection/classification method by designing a CNN signature representation that is revealing and easy to compute. It demonstrates a significant performance improvement over the existing competing methods with more general assumptions about the Trojan/backdoor attacks.\\
 
 
\noindent{\bf Acknowledgment} This work is supported by NSF IIS-1717431
and NSF IIS-1618477. We thank Rajesh Gupta and 
Mani Srivastava for valuable discussions.

\bibliographystyle{splncs04}
\bibliography{egbib}
\end{document}